\documentclass[11pt,letterpaper]{article}
\usepackage{emnlp2017}
\usepackage{times}
\usepackage{latexsym}

\usepackage{url}

\usepackage[utf8]{inputenc} % allow utf-8 input
\usepackage[T1]{fontenc}    % use 8-bit T1 fonts
\usepackage{hyperref}       % hyperlinks
\usepackage{url}            % simple URL typesetting
\usepackage{booktabs}       % professional-quality tables
\usepackage{amsfonts}       % blackboard math symbols
\usepackage{nicefrac}       % compact symbols for 1/2, etc.
\usepackage{microtype}      % microtypography

\usepackage{amsmath}
\usepackage{amsfonts}
\usepackage{amssymb}
\usepackage{wrapfig}
\usepackage{bm}
\usepackage{subcaption}
\usepackage{psfrag}
\usepackage{tikz}
\usepackage{epstopdf}

\usepackage{color}
\usepackage{tikz}
\usetikzlibrary{arrows,shapes,snakes,automata,backgrounds,fit,petri}
\usepackage{adjustbox}

\usepackage{algorithm}
\usepackage{algorithmic}

\newcommand{\beq}{\vspace{0mm}\begin{equation}}
\newcommand{\eeq}{\vspace{0mm}\end{equation}}
\newcommand{\beqs}{\vspace{0mm}\begin{eqnarray}}
\newcommand{\eeqs}{\vspace{0mm}\end{eqnarray}}
\newcommand{\barr}{\begin{array}}
\newcommand{\earr}{\end{array}}

\newcommand{\Vmat}[0]{{{\bf V}}}
\newcommand{\Wmat}[0]{{{\bf W}}}
\newcommand{\Xmat}[0]{{{\bf X}}}

\newcommand{\bv}[0]{{\boldsymbol{b}}}
\newcommand{\cv}[0]{{\boldsymbol{c}}}

\newcommand{\hv}[0]{{\boldsymbol{h}}}

\newcommand{\xv}{\boldsymbol{x}}
\newcommand{\yv}{\boldsymbol{y}}
\newcommand{\zv}{\boldsymbol{z}}

\newcommand{\R}{\mathbb{R}}

\newcommand{\Hcal}{\mathcal{H}}
\newcommand{\Vcal}{\mathcal{V}}

\emnlpfinalcopy

\def\emnlppaperid{***}

\title{Learning Generic Sentence Representations Using \\ Convolutional Neural Networks}

\author{
	Zhe Gan$^\dag$, Yunchen Pu$^\dag$, Ricardo Henao$^\dag$, Chunyuan Li$^\dag$, Xiaodong He$^\ddag$, Lawrence Carin$^\dag$\\
	$^\dag$Duke University, $^\ddag$Microsoft Research, Redmond, WA 98052, USA \\ \texttt{\{zg27, yp42, r.henao, cl319, lcarin\}@duke.edu} \\
	\texttt{xiaohe@microsoft.com}
}

\date{}

\begin{document}

\maketitle

\begin{abstract}
  We propose a new encoder-decoder approach to learn distributed sentence representations that are applicable to multiple purposes. The model is learned by using a
  convolutional neural network as an encoder to map an input sentence into a continuous vector, and using a long short-term memory recurrent neural network as a decoder. Several tasks are considered, including sentence reconstruction and future sentence prediction. Further, a hierarchical encoder-decoder model is proposed to encode a sentence to
  predict multiple future sentences.
  By training our models on a large collection of novels, we obtain a highly generic convolutional sentence encoder that performs well in practice. Experimental results on several benchmark datasets, and across a broad range of applications, demonstrate the superiority of the proposed model over competing methods.
\end{abstract}

\section{Introduction}

Learning sentence representations is central to many natural language modeling applications. The aim of a model for this task is to learn fixed-length feature vectors that encode the semantic and syntactic properties of sentences. Deep learning techniques have shown promising performance on sentence modeling, via feedforward neural networks~\citep{huang2013learning}, recurrent neural networks (RNNs)~\citep{hochreiter1997long}, convolutional neural networks (CNNs)~\citep{kalchbrenner2014convolutional,kim2014convolutional,shen2014latent}, and recursive neural networks~\citep{socher2013recursive}. 
Most of these models are \emph{task-dependent}: they are trained specifically for a certain task. However, these methods may become inefficient when we need to repeatedly learn sentence representations for a large number of different tasks, because they may require retraining a new model for each individual task. In this paper, in contrast, we are primarily interested in learning \emph{generic} sentence representations that can be used across domains. 

Several approaches have been proposed for learning generic sentence embeddings. The paragraph-vector model of~\citet{le2014distributed} incorporates a global context vector into the log-linear neural language model~\citep{mikolov2013distributed} to learn the sentence representation; however, at prediction time, one needs to perform gradient descent to compute a new vector. The sequence autoencoder of~\citet{dai2015semi} describes an encoder-decoder model to reconstruct the input sentence, while the skip-thought model of~\citet{kiros2015skip} extends the encoder-decoder model to reconstruct the surrounding sentences of an input sentence. Both the encoder and decoder of the methods above are modeled as RNNs.

\begin{figure*}
	\centering
	\includegraphics[width=0.99\textwidth]{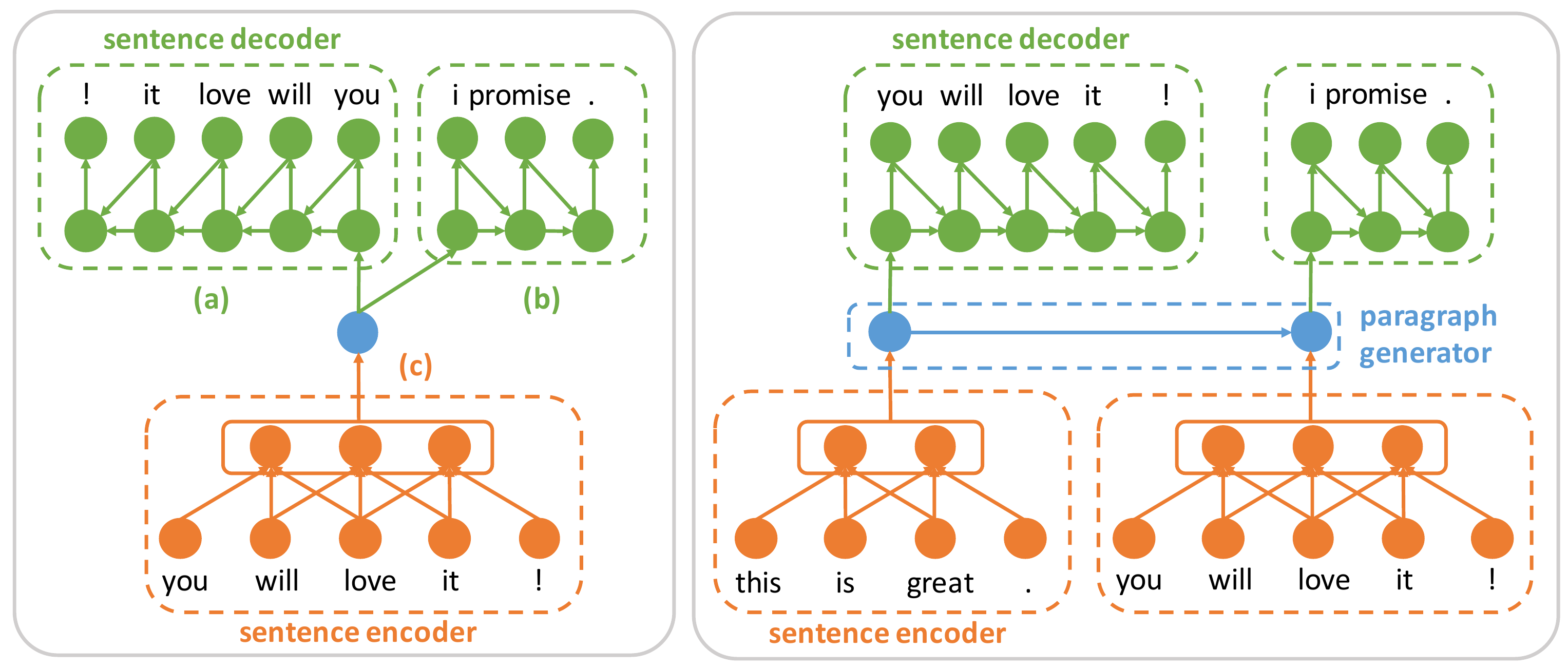}
	\caption{Illustration of the CNN-LSTM encoder-decoder models. The sentence encoder is a CNN, the sentence decoder is an LSTM, and the paragraph generator is another LSTM. (Left) (a)+(c) represents the autoencoder; (b)+(c) represents the future predictor; (a)+(b)+(c) represents the composite model. (Right) hierarchical model. In this example, the input contiguous sentences are: \emph{this is great. you will love it! i promise}.}
	\label{fig:framework}
\end{figure*}
CNNs have recently achieved excellent results in various task-dependent natural language applications as the sentence encoder~\citep{kalchbrenner2014convolutional,kim2014convolutional,hu2014convolutional}. This motivates us to propose a CNN encoder for learning generic sentence representations within the framework of encoder-decoder models proposed by~\citet{sutskever2014sequence,cho2014learning}. Specifically, a CNN encoder performs convolution and pooling operations on an input sentence, then uses a fully-connected layer to produce a fixed-length encoding of the sentence. This encoding vector is then fed into a long short-term memory (LSTM) recurrent network to produce a target sentence. Depending on the task, we propose three models: (\emph{i}) \emph{CNN-LSTM autoencoder}: this model seeks to reconstruct the original input sentence, by capturing the \emph{intra}-sentence information; (\emph{ii}) \emph{CNN-LSTM future predictor}: this model aims to predict a future sentence, by leveraging \emph{inter}-sentence information; (\emph{iii}) \emph{CNN-LSTM composite model}: in this case, there are two LSTMs, decoding the representation to the input sentence itself and a future sentence.
This composite model aims to learn a sentence encoder that captures both \emph{intra-} and \emph{inter-}sentence information.

The proposed CNN-LSTM future predictor model only considers the immediately subsequent sentence as context. In order to capture longer-term dependencies between sentences, we further introduce a hierarchical encoder-decoder model. This model abstracts the RNN language model of~\citet{mikolov2010recurrent} to the sentence level. That is, instead of using the current word in a sentence to predict future words (sentence continuation), we encode a sentence to predict multiple future sentences (paragraph continuation). This model is termed \emph{hierarchical CNN-LSTM model}.

As in~\citet{kiros2015skip}, we first train our proposed models on a large collection of novels. We then evaluate the CNN sentence encoder as a generic feature extractor for 8 tasks: semantic relatedness, paraphrase detection, image-sentence ranking and 5 standard classification benchmarks. In these experiments, we train a linear classifier on top of the extracted sentence features, without additional fine-tuning of the CNN. We show that our trained sentence encoder yields generic representations that perform as well as, or better, than those of~\citet{kiros2015skip,hill2016learning}, in all the tasks considered. 

Summarizing, the main contribution of this paper is a new class of CNN-LSTM encoder-decoder models that is able to leverage the vast quantity of unlabeled text for learning generic sentence representations. Inspired by the skip-thought model~\cite{kiros2015skip}, we have further explored different variants: 
(\emph{i}) CNN is used as the sentence encoder rather than RNN; (\emph{ii}) larger context windows are considered:
we propose the hierarchical CNN-LSTM model to encode a sentence for predicting multiple future sentences.

\section{Model description}
\subsection{CNN-LSTM model}\label{sec:cnn_lstm}
Consider the sentence pair $(s_x,s_y)$. The encoder, a CNN, encodes the first sentence $s_x$ into a feature vector $\zv$, which is then fed into an LSTM decoder that predicts the second sentence $s_y$. 
Let $w_x^t \in \{1,\dots,V\}$ represent the $t$-th word in sentences $s_x$, where $w_x^t$ indexes one element in a $V$-dimensional set (vocabulary); $w_y^t$ is defined similarly w.r.t. $s_y$.
Each word $w_x^t$ is embedded into a $k$-dimensional vector $\xv_t=\Wmat_e[w_x^t]$, where $\Wmat_e \in \R^{k\times V}$ is a word embedding matrix (learned), and notation $\Wmat_e[v]$ denotes the $v$-th column of matrix $\Wmat_e$. Similarly, we let $\yv_t=\Wmat_e[w_y^t]$. 

\paragraph{CNN encoder} 
The CNN architecture in~\citet{kim2014convolutional,collobert2011natural} is used for sentence encoding, which consists of a convolution layer and a max-pooling operation over the entire sentence for each feature map. A sentence of length $T$ (padded where necessary) is represented as a matrix $\Xmat \in \R^{k\times T}$, by concatenating its word embeddings as columns, \emph{i.e.}, the $t$-th column of $\Xmat$ is $\xv_t$.

A convolution operation involves a filter $\Wmat_c \in \R^{k\times h}$, applied to a window of $h$ words to produce a new feature. According to~\citet{collobert2011natural}, we can induce one feature map $\cv=f(\Xmat \ast \Wmat_c +\bv) \in \R^{T-h+1}$, where $f(\cdot)$ is a nonlinear activation function such as the hyperbolic tangent used in our experiments, $\bv\in\R^{T-h+1}$ is a bias vector, and $\ast$ denotes the convolutional operator. Convolving the same filter with the $h$-gram at every position in the sentence allows the features to be extracted independently of their position in the sentence. We then apply a max-over-time pooling operation~\cite{collobert2011natural} to the feature map and take its maximum value, {\em i.e.}, $\hat{c}=\max\{\cv\}$, as the feature corresponding to this particular filter. This pooling scheme tries to capture the most important feature, \emph{i.e.}, the one with the highest value, for each feature map, effectively filtering out less informative compositions of words. Further, pooling also guarantees that the extracted features are independent of the length of the input sentence.

The above process describes how one feature is extracted from one filter. In practice, the model uses multiple filters with varying window sizes~\cite{kim2014convolutional}. Each filter can be considered as a linguistic feature detector that learns to recognize a specific class of $n$-grams (or $h$-grams, in the above notation). 
However, since the $h$-grams are computed in the embedding space, the model naturally handles similar $h$-grams composed of synonyms. 
Assume we have $m$ window sizes, and for each window size, we use $d$ filters; then we obtain a $md$-dimensional vector to represent a sentence. 

Compared with the LSTM encoders used in~\citet{kiros2015skip,dai2015semi,hill2016learning}, a CNN encoder may have the following advantages. First, the sparse connectivity of a CNN, which indicates fewer parameters are required, typically improves its statistical efficiency as well as reduces memory requirements~\cite{Goodfellow2016Book}. For example, excluding the number of parameters used in the word embeddings, our trained CNN sentence encoder has 3 million parameters, while the skip-thought vector of~\citet{kiros2015skip} contains 40 million parameters. 
Second, a CNN is easy to implement in parallel over the whole sentence, while an LSTM needs sequential computation. 

\paragraph{LSTM decoder}
The CNN encoder maps sentence $s_x$ into a vector $\zv$. 
The probability of a length-$T$ sentence $s_y$ given the encoded feature vector $\zv$ is defined as
\begin{align}\label{eq:lik}
	p(s_y|\zv)=\prod_{t=1}^{T}p(w_y^t|w_y^{0},\ldots,w_y^{t-1},\zv) 
	%p(s_y|\zv)=\textstyle{\prod_{t=1}^{T}}p(w_y^t|w_y^{0},\ldots,w_y^{t-1},\zv) 
\end{align}
where $w_y^0$ is defined as a special start-of-the-sentence token.
All the words in the sentence are sequentially generated using the RNN, until the end-of-the-sentence symbol is generated. Specifically,
each conditional $p(w_y^t|w_y^{<t},\zv)$, where $<t=\{0,\ldots,t-1\}$, is specified as $\mbox{softmax} (\Vmat \hv_t)$, where $\hv_t$, the hidden units, are recursively updated through $\hv_t = \Hcal(\yv_{t-1},\hv_{t-1},\zv)$, and $\hv_0$ is defined as a zero vector ($\hv_0$ is not updated during training). 
$\Vmat$ is a weight matrix used for computing a distribution over words. Bias terms are omitted for simplicity throughout the paper. 
The transition function $\Hcal(\cdot)$ is implemented with an LSTM~\cite{hochreiter1997long}. 

Given the sentence pair $(s_x,s_y)$, the objective function is the sum of the log-probabilities of the target sentence conditioned on the encoder representation in \eqref{eq:lik}: $\sum_{t=1}^{T} \log p(w_y^t|w_y^{<t},\zv)$. The total objective is the above objective summed over all the sentence pairs.

\paragraph{Applications}
Inspired by~\citet{srivastava2015unsupervised}, we propose three models: (\emph{i}) an autoencoder, (\emph{ii}) a future predictor, and (\emph{iii}) the composite model. These models share the same CNN-LSTM model architecture, but are different in terms of the choices of the target sentence. An illustration of the proposed encoder-decoder models is shown in Figure~\ref{fig:framework}(left).

The autoencoder ($i$) aims to reconstruct the same sentence as the input. The intuition behind this is that an autoencoder learns to represent the data using features that explain its own important factors of variation, and hence model the internal structure of sentences, effectively capturing the \emph{intra}-sentence information. Another natural task is encoding an input sentence to predict the subsequent sentence. The future predictor ($ii$) achieves this, effectively capturing the \emph{inter}-sentence information, which has been shown to be useful to learn the semantics of a sentence~\cite{kiros2015skip}. These two tasks can be combined to create a composite model ($iii$), where the CNN encoder is asked to learn a feature vector that is useful to simultaneously reconstruct the input sentence and predict a future sentence. This composite model encourages the sentence encoder to incorporate contextual information both within and beyond the sentence. 

\subsection{Hierarchical CNN-LSTM model}\label{sec:hier_cnn_lstm}
\begin{figure}[t!]
	\centering
	\hspace{-3mm}
	\includegraphics[width=1.0\linewidth]{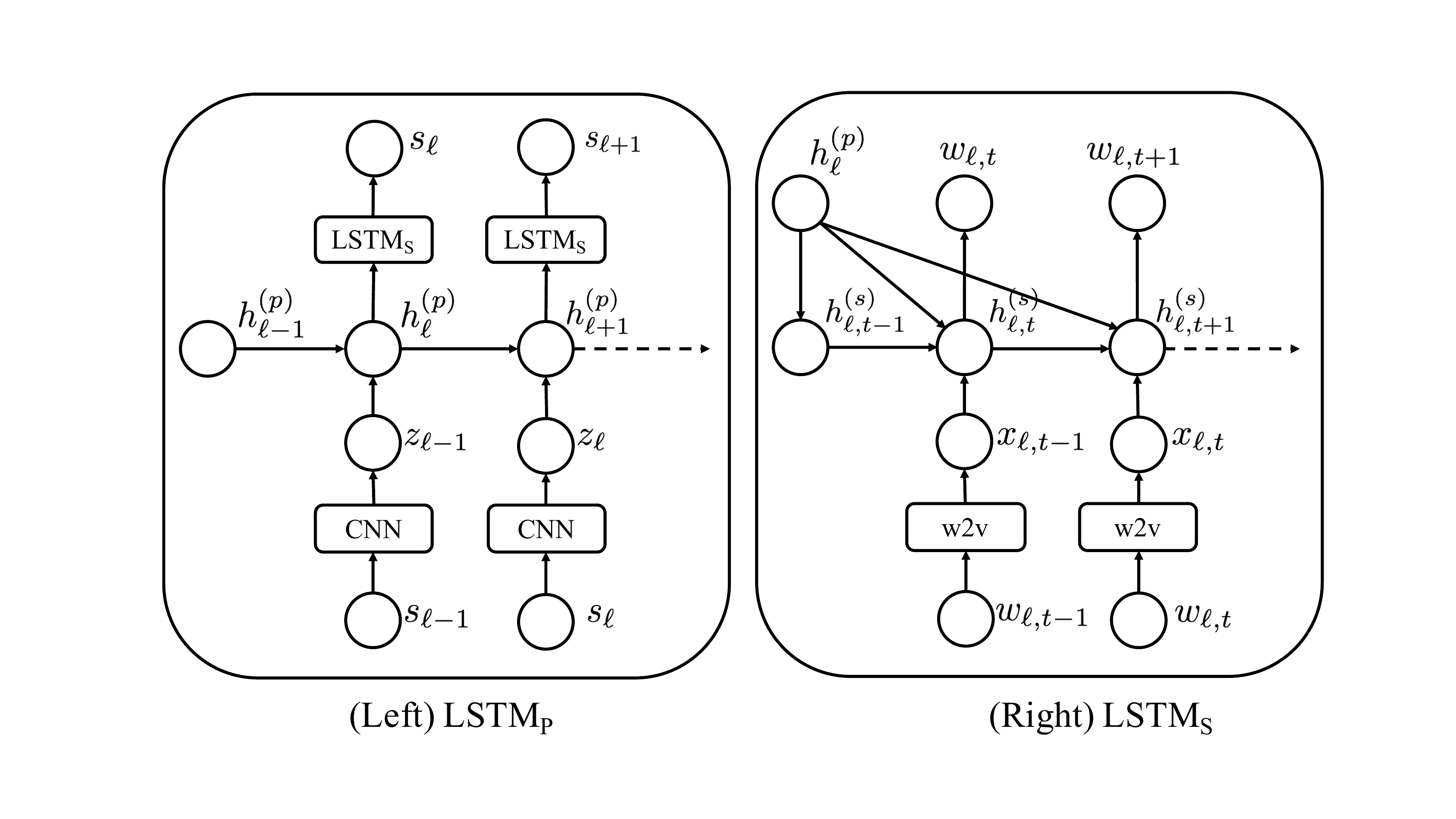} 
	\caption{{Detailed illustration of the hierarchical CNN-LSTM model. (Left) LSTM paragraph generator. (Right) LSTM sentence decoder.}}
	\label{fig:hier_model}
	%\vspace{-2.0mm}
\end{figure}

The future predictor described in Section~\ref{sec:cnn_lstm} only considers the immediately subsequent sentence as context. By utilizing a larger surrounding context, it is likely that we can learn even higher-quality sentence representations. Inspired by the standard RNN-based language model~\cite{mikolov2010recurrent} that uses the current word to predict future words, we propose a hierarchical encoder-decoder model that encodes the current sentence to predict multiple future sentences. An illustration of the hierarchical model is shown in Figure~\ref{fig:framework}(right), with details provided in Figure~\ref{fig:hier_model}. 

Our proposed hierarchical model characterizes the hierarchy {\em word-sentence-paragraph}. A paragraph is modeled as a sequence of sentences, and each sentence is modeled as a sequence of words. Specifically, assume we are given a paragraph $D=(s_1,\ldots,s_L)$, that consists of $L$ sentences. The probability for paragraph $D$ is then defined as
\begin{align}
	p(D) = \prod_{\ell=1}^L p(s_{\ell}|s_{<\ell})  \label{eqn:hier_model_p}
	%p(D) = \textstyle{\prod_{\ell=1}^L} p(s_{\ell}|s_{<\ell})  \label{eqn:hier_model_p}
\end{align}
where $s_0$ is defined as a special start-of-the-paragraph token.
As shown in Figure~\ref{fig:hier_model}(left), each $p(s_\ell|s_{<\ell})$ in~(\ref{eqn:hier_model_p}) is calculated as
\begin{align}
	p(s_\ell|s_{<\ell}) &= p(s_{\ell}|\hv_\ell^{(p)}) \label{eqn:lstm_s}\\
	\hv_{\ell}^{(p)} &= \mbox{LSTM}_p (\hv_{\ell-1}^{(p)},\zv_{\ell-1}) \label{eqn:lstm_p} \\ 
	\zv_{\ell-1} &= \mbox{CNN}(s_{\ell-1}) \label{eqn:cnn}
\end{align}
where $\hv_{\ell}^{(p)}$ denotes the $\ell$-th hidden state of the LSTM paragraph generator, and $\hv_0^{(p)}$ is fixed as a zero vector.
The CNN in~(\ref{eqn:cnn}) is as described in Section~\ref{sec:cnn_lstm}, encoding the sentence $s_{\ell-1}$ into a vector representation $\zv_{\ell-1}$. 

Equation (\ref{eqn:lstm_p}) serves as the paragraph-level language model~\cite{mikolov2010recurrent}, which encodes all the previous sentence representations $\zv_{<\ell}$ into a vector representation $\hv_{\ell}^{(p)}$. This hidden state $\hv_{\ell}^{(p)}$ is used to guide the generation of the $\ell$-th sentence through the decoder (\ref{eqn:lstm_s}), which is defined as
\begin{align}
	p(s_{\ell}|\hv_\ell^{(p)}) = \prod_{t=1}^{T_{\ell}} p(w_{\ell,t}|w_{\ell,<t}, \hv_\ell^{(p)}) \label{eqn:hier_model_s}
	%p(s_{\ell}|\hv_\ell^{(p)}) = \textstyle{\prod_{t=1}^{T_{\ell}}} p(w_{\ell,t}|w_{\ell,<t}, \hv_\ell^{(p)}) \label{eqn:hier_model_s}
\end{align}
where $w_{\ell,0}$ is defined as a special start-of-the-sentence token. $T_{\ell}$ is the length of sentence $\ell$, and $w_{\ell,t}$ denotes the $t$-th word in sentence $\ell$.  
As shown in Figure~\ref{fig:hier_model}(right), each $p(w_{\ell,t}|w_{\ell,<t},\hv_\ell^{(p)})$  in~(\ref{eqn:hier_model_s}) is calculated as
\begin{align}
	p(w_{\ell,t}|w_{\ell,<t},\hv_\ell^{(p)}) &= \mbox{softmax}(\Vmat \hv_{\ell,t}^{(s)}) \\
	\hv_{\ell,t}^{(s)} = \mbox{LSTM}_s &(\hv_{\ell,t-1}^{(s)},\xv_{\ell,t-1},\hv_{\ell}^{(p)})
\end{align}
where $\hv_{\ell,t}^{(s)}$ denotes the $t$-th hidden state of the LSTM decoder for sentence $\ell$, $\xv_{\ell,t-1}$ denotes the word embedding for $w_{\ell,t-1}$, and
$\hv_{\ell,0}^{(s)}$ is fixed as a zero vector for all $\ell=1,\ldots,L$.  $\Vmat$ is a weight matrix used for computing distribution over words.

\section{Related work}

Various methods have been proposed for sentence modeling, which generally fall into two categories. The first consists of models trained specifically for a certain task, typically combined with downstream applications. Several models have been proposed along this line, ranging from simple additional composition of the word vectors~\cite{mitchell2010composition,yu2015learning,iyyer2015deep} to those based on complex nonlinear functions like recursive neural networks~\cite{socher2011dynamic,socher2013recursive}, convolutional neural networks~\cite{kalchbrenner2014convolutional,hu2014convolutional,johnson2014effective,zhang2015character,gan2017character}, and recurrent neural networks~\cite{tai2015improved,lin2017structured}.

The other category consists of methods aiming to learn generic sentence representations that can be used across domains.
This includes the paragraph vector~\cite{le2014distributed}, skip-thought vector~\cite{kiros2015skip}, and the sequential denoising autoencoders~\cite{hill2016learning}. 
~\citet{hill2016learning} also proposed a sentence-level log-linear bag-of-words
(BoW) model, where a BoW representation of an input sentence is used to predict adjacent
sentences that are also represented as BoW. Most recently,~\citet{wieting2015towards,Arora2017simple,Pagliardini2017unsupervised} proposed methods in which sentences are represented as a weighted average of fixed (pre-trained) word vectors. 
Our model falls into this category, and is most related to~\citet{kiros2015skip}.

However, there are two key aspects that make our model different from~\citet{kiros2015skip}. First, we use CNN as the sentence encoder.  The combination of CNN and LSTM has been considered in image captioning~\cite{karpathy2015deep}, and in some recent work on machine translation~\cite{kalchbrenner2013recurrent,meng2015encoding,gehring2016convolutional}. Our utilization of a CNN is different, and more importantly, the ultimate goal of our model is different. Our work aims to use a CNN to learn generic sentence embeddings.

Second, we use the hierarchical CNN-LSTM model to predict multiple future sentences, rather than the surrounding two sentences as in~\citet{kiros2015skip}. 
Utilizing a larger context window aids our model to learn better sentence representations, capturing longer-term dependencies between sentences.
Similar work to this hierarchical language modeling can be found in~\citet{li2015hierarchical,sordoni2015hierarchical,lin2015hierarchical,wanglarger}. Specifically,~\citet{li2015hierarchical,sordoni2015hierarchical} uses an LSTM for the sentence encoder, while~\citet{lin2015hierarchical} uses a bag-of-words to represent sentences.

\section{Experiments}

\begin{table*}[t!]
	\centering
	\small
	\begin{tabular}{c|l|l|l|l}
		\toprule
		\textbf{A} & you needed me? & this is great. & its lovely to see you. & he had thought he was going crazy.\\
		\textbf{B} & you got me? & this is awesome. & its great to meet you. & i felt like i was going crazy.\\
		\textbf{C} & i got you. & you are awesome. & its great to meet him. & i felt like to say the right thing.\\
		\midrule
		\textbf{D} & i needed you. & you are great. & its lovely to see him. & he had thought to say the right thing. \\
		\bottomrule
	\end{tabular}
	\caption{Vector ``compositionality'' using element-wise addition and subtraction. Let $\zv(s)$ denote the vector representation $\zv$ of a given sentence $s$. We first calculate  $\zv^\star$=$\zv$(A)-$\zv$(B)+$\zv$(C). The resulting vector is then sent to the LSTM to generate sentence D.} \label{Table:sentence_reasoning}
\end{table*}

\begin{table*}[h!]
	\centering
	\small
	\begin{tabular}{l}
		\toprule
		\textbf{Query and nearest sentence}  \\
		\midrule
		johnny nodded his curly head , and then his breath eased into an even rhythm . \\
		aiden looked at my face for a second , and then his eyes trailed to my extended hand . \\
		\midrule
		i yelled in frustration , throwing my hands in the air . \\
		i stand up , holding my hands in the air .  \\
		\midrule
		i loved sydney , but i was feeling all sorts of homesickness . \\
		i loved timmy , but i thought i was a self-sufficient person . \\
		\midrule
		`` i brought sad news to mistress betty , '' he said quickly , taking back his hand . \\
		`` i really appreciate you taking care of lilly for me , '' he said sincerely , handing me the money . \\
		\midrule
		`` i am going to tell you a secret , '' she said quietly , and he leaned closer .\\
		`` you are very beautiful , '' he said , and he leaned in . \\
		\midrule
		she kept glancing out the window at every sound , hoping it was jackson coming back . \\
		i kept checking the time every few minutes , hoping it would be five oclock . \\
		\midrule
		leaning forward , he rested his elbows on his knees and let his hands dangle between his legs .  \\
		stepping forward , i slid my arms around his neck and then pressed my body flush against his . \\
		\midrule
		i take tris 's hand and lead her to the other side of the car , so we can watch the city disappear behind us .   \\
		i take emma 's hand and lead her to the first taxi , everyone else taking the two remaining cars .  \\
		\bottomrule
	\end{tabular}
	\caption{Query-retrieval examples. In each case (block of rows), the first sentence is a query, while the second sentence is the retrieved result from a random subset of 1 million sentences from the BookCorpus dataset.}\label{Table:sentence_retrieval}
\end{table*}

We first provide qualitative analysis of our CNN encoder, and then present experimental results on 8 tasks: 5 classification benchmarks, paraphrase detection, semantic relatedness and image-sentence ranking. As in~\citet{kiros2015skip}, we evaluate the capabilities of our encoder as a generic feature extractor. To further demonstrate the advantage of our learned generic sentence representations, we also fine-tune our trained sentence encoder on the 5 classification benchmarks. All the CNN-LSTM models are trained using the BookCorpus dataset~\cite{zhu2015aligning}, which consists of 70 million sentences from over 7000 books.

We train four models in total: (\emph{i}) an autoencoder, (\emph{ii}) a future predictor, (\emph{iii}) the composite model, and (\emph{iv}) the hierarchical  model. For the CNN encoder, we employ filter windows ($h$) of sizes \{3,4,5\} with 800 feature maps each, hence each sentence is represented as a 2400-dimensional vector. For both, the LSTM sentence decoder and paragraph generator, we use one hidden layer of 600 units.

The CNN-LSTM models are trained with a vocabulary size of 22,154 words. In order to learn a generic sentence encoder that can encode a large number of possible words, we use two methods of considering words not in the training set. 
Suppose we have a large pretrained word embedding matrix, such as the publicly available \emph{word2vec} vectors~\cite{mikolov2013distributed}, in which all test words are assumed to reside. 

The first method learns a linear mapping between the \emph{word2vec} embedding space $\Vcal_{w2v}$ and the learned word embedding space $\Vcal_{cnn}$ by solving a linear regression problem~\citep{kiros2015skip}. Thus, any word from $\Vcal_{w2v}$ can be mapped into $\Vcal_{cnn}$ for encoding sentences. 
The second method fixes the word vectors in $\Vcal_{cnn}$ as the corresponding word vectors in $\Vcal_{w2v}$ , and we do not update the word embedding parameters during training. Thus, any word vector from $\Vcal_{w2v}$ can be naturally used to encode sentences. 
By doing this, our trained sentence encoder can successfully encode 931,331 words.

For training, all weights in the CNN and non-recurrent weights in the LSTM are initialized from a uniform distribution in [-0.01,0.01]. Orthogonal initialization is employed on the recurrent matrices in the LSTM. All bias terms are initialized to zero. The initial forget gate bias for LSTM is set to 3.  Gradients are clipped if the norm of the parameter vector exceeds 5~\cite{sutskever2014sequence}. The Adam algorithm~\cite{kingma2014adam} with learning rate $2\times10^{-4}$ is utilized for optimization. For all the CNN-LSTM models, we use mini-batches of size 64. For the hierarchical CNN-LSTM model, we use mini-batches of size 8, and each paragraph is composed of 8 sentences. We do not perform any regularization other than dropout~\cite{srivastava2014dropout}. All experiments are implemented in Theano~\cite{bastien2012theano}, using a NVIDIA GeForce GTX TITAN X GPU with 12GB memory. 

\subsection{Qualitative analysis}
We first demonstrate that the sentence representation learned by our model exhibits a structure that makes it possible to perform analogical reasoning using simple vector arithmetics, as illustrated in Table~\ref{Table:sentence_reasoning}. It demonstrates that the arithmetic operations on the sentence representations correspond to word-level addition and subtractions. 
For instance, in the 3rd example, our encoder captures that the difference between sentence B and C is \emph{``you"} and \emph{``him"}, so that the former word in sentence A is replaced by the latter ({\em i.e.}, \emph{``you''}-\emph{``you''}+\emph{``him''}=\emph{``him''}), resulting in sentence D.

Table~\ref{Table:sentence_retrieval} shows nearest neighbors of sentences from a CNN-LSTM autoencoder trained on the BookCorpus dataset. Nearest neighbors are scored by cosine similarity from a random sample of 1 million sentences from the BookCorpus dataset. As can be seen, our encoder learns to accurately capture semantic and syntax of the sentences.
\subsection{Quantitative evaluations}
\begin{table*}[t!]
	\centering
	\small
	\begin{tabular}{lcccccc}
		\toprule
		\textbf{Method} & \textbf{MR} & \textbf{CR} & \textbf{SUBJ} & \textbf{MPQA} & \textbf{TREC} &  \textbf{MSRP(Acc/F1)} \\
		\midrule
		ParagraphVec DM~\cite{hill2016learning} & 61.5 & 68.6 & 76.4 & 78.1 & 55.8  & 73.6 / 81.9\\
		%\hline
		SDAE~\cite{hill2016learning} & 67.6 & 74.0 & 89.3 & 81.3& 77.6  & 76.4 / 83.4\\
		%\hline
		SDAE+emb.~\cite{hill2016learning} & 74.6 & 78.0 & 90.8 & 86.9 & 78.4  & 73.7 / 80.7\\
		FastSent~\cite{hill2016learning} & 70.8 & 78.4 & 88.7 & 80.6 & 76.8  & 72.2 / 80.3\\
		\midrule 
		uni-skip~\cite{kiros2015skip}  & 75.5 & 79.3 & 92.1 & 86.9 & 91.4  & 73.0 / 81.9 \\
		bi-skip~\cite{kiros2015skip}  & 73.9 & 77.9 & 92.5 & 83.3 & 89.4  & 71.2 / 81.2\\
		combine-skip~\cite{kiros2015skip} & 76.5 & 80.1 & 93.6 & 87.1 & 92.2 & 73.0 / 82.0\\
		\midrule
		\emph{Our Results$^\dagger$} & & & & & & \\
		\midrule
		autoencoder   & 75.53 & 78.97 & 91.97 & 87.96 & 89.8 & 73.61 / 82.14\\
		future predictor  & 72.56 & 78.44 & 90.72 & 87.48 & 86.6 & 71.87 / 81.68 \\
		hierarchical model  & 75.20 & 77.99 & 91.66 & 88.21 & 90.0 & 73.96 / 82.54\\
		composite model  & 76.34 & 79.93 & 92.45 & 88.77 & 91.4 & 74.65 / 82.21\\
		combine$^\ddagger$  & 77.21 & 80.85 & 93.11 & 89.09 & 91.8 & 75.52 / 82.62 \\
		\midrule
		hierarchical model+emb.  & 75.30 & 79.37 & 91.94  & 88.48  & 90.4  & 74.25 / 82.70 \\
		composite model+emb.  & 77.16 & 80.64 & 92.14 & 88.67 & 91.2 & 74.88 / 82.28 \\
		combine+emb.$^\ddagger$  & {\bf 77.77} & {\bf 82.05}  & {\bf 93.63} & {\bf 89.36} & {\bf92.6}  & {\bf 76.45} / {\bf 83.76} \\
		\midrule
		\emph{Task-dependent methods} & & & & & & \\
		\midrule
		CNN~\cite{kim2014convolutional} & 81.5 & 85.0 & 93.4 & 89.6 & 93.6 & $-$  \\
		AdaSent~\cite{zhao2015self} & 83.1 & 86.3 & 95.5 & 93.3 & 92.4 & $-$  \\
		Bi-CNN-MI~\cite{yin2015convolutional} & $-$ & $-$ & $-$ & $-$ & $-$ & 78.1/84.4  \\
		MPSSM-CNN~\cite{he2015multi} & $-$ & $-$ & $-$ & $-$ & $-$ & 78.6/84.7 \\
		\bottomrule
	\end{tabular}
	\caption{{Classification accuracies on several standard benchmarks. The last column shows results on the task of paraphrase detection, where the evaluation metrics are classification accuracy and F1 score. $^\dagger$The first and second block in our results were obtained using the first and second method of considering words not in the training set, respectively. $^\ddagger$``combine'' means concatenating the feature vectors learned from both the hierarchical model and the composite model. }}\label{Table:classification_result}
	%\vspace{-3.5mm}
\end{table*}

\paragraph{Classification benchmarks}
We first study the task of sentence classification on 5 datasets: \emph{MR}~\cite{pang2005seeing}, \emph{CR}~\cite{hu2004mining}, \emph{SUBJ}~\cite{pang2004sentimental},  \emph{MPQA}~\cite{wiebe2005annotating} and \emph{TREC}~\cite{li2002learning}. On all the datasets, we separately train a logistic regression model on top of the extracted sentence features. We restrict our comparison to methods that also aims to learn generic sentence embeddings for fair comparison. We also provide the state-of-the-art results using task-dependent learning methods for reference. Results are summarized in Table~\ref{Table:classification_result}. Our CNN encoder provides better results than the combine-skip model of~\citet{kiros2015skip} on all the 5 datasets. 

We highlight some observations. First, the autoencoder performs better than the future predictor, indicating that the \emph{intra-}sentence information may be more important for classification than the \emph{inter-}sentence information. Second, the hierarchical model performs better than the future predictor, demonstrating the importance of capturing long-term dependencies across \emph{multiple} sentences. Our combined model, which concatenates the feature vectors learned from both the hierarchical model and the composite model, performs the best. This may be due to that: (\emph{i}) both \emph{intra-} and long-term \emph{inter-}sentence information are leveraged; (\emph{ii}) it is easier to linearly separate the feature vectors in higher dimensional spaces. 
Further, using (fixed) pre-trained word embeddings consistently provides better performance than using the learned word embeddings. This may be due to that \emph{word2vec} provides more generic word representations, since it is trained on the large Google News dataset (containing 100 billion words)~\cite{mikolov2013distributed}.

\begin{figure}[t!]
	\centering
	\hspace{-3mm}
	\includegraphics[width=0.5\linewidth]{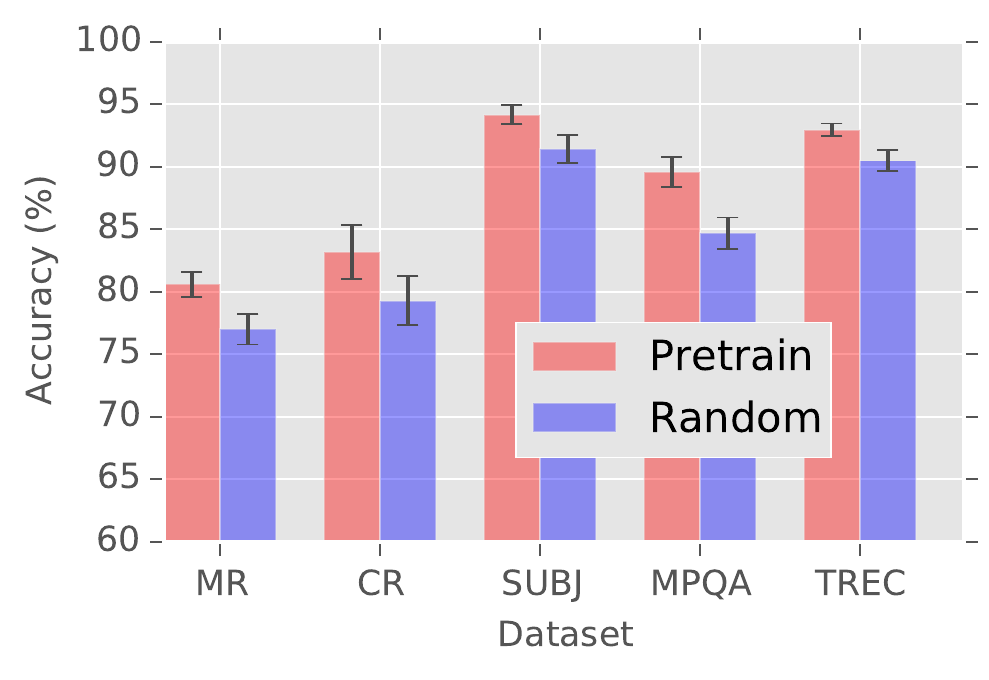} 
	\includegraphics[width=0.5\linewidth]{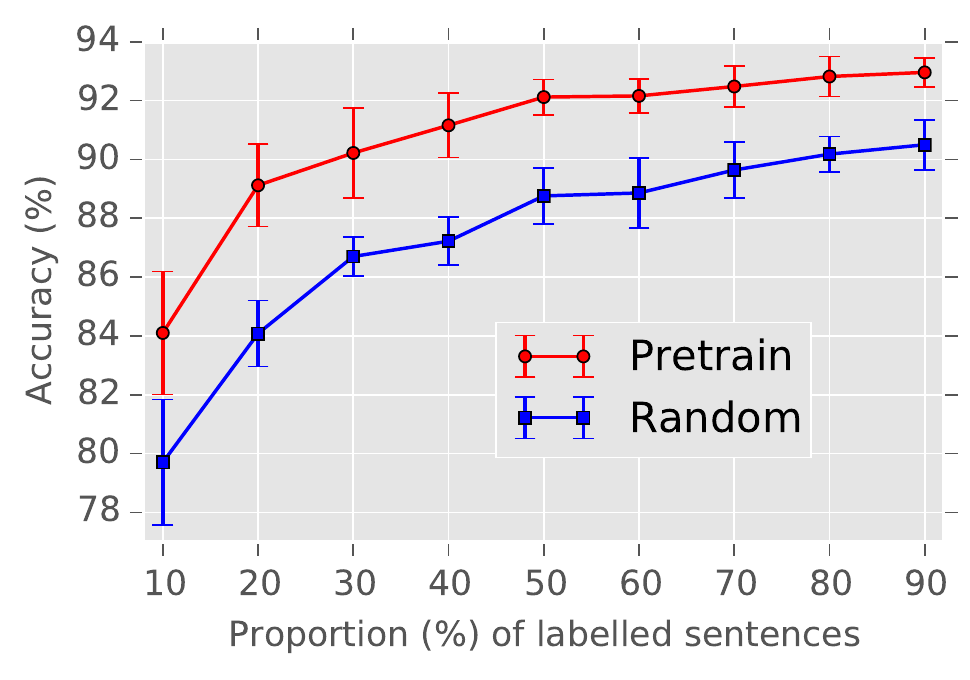} 
	\caption{{(Left) Effect of pretraining on the 5 classification benchmarks. The error bars are over 10 different runs. (Right) Effect of pretraining on accuracy for the TREC dataset, in terms of change in the size of the labeled training set. The error bars are over 10 different samples of training sets. Pretraining means initializing the CNN parameters from the trained CNN-LSTM composite model.}}
	\label{fig:pretrain_vs_random}
	%\vspace{-4.0mm}
\end{figure}

To further demonstrate the advantage of the learned generic representations, we train a CNN classifier (\emph{i.e.}, a CNN encoder with a logistic regression model on top) with two different initialization strategies: random initialization and initialization with trained parameters from the CNN-LSTM composite model. Results are shown in Figure~\ref{fig:pretrain_vs_random}(left). The pretraining provides substantial improvements (3.52\% on average) over random initialization of CNN parameters. Figure~\ref{fig:pretrain_vs_random}(right) shows the effect of pretraining as the number of labeled sentences is varied. For the TREC dataset, the performance improves from 79.7\% to 84.1\% when only 10\% sentences are labeled. As the size of the set of labeled sentences grows, the improvement becomes smaller, as expected.
For future work, our CNN-LSTM model can be also used for semi-supervised learning, with the autoencoder on all the data (labeled and unlabled), and the classifier only on the labeled data. 

\paragraph{Paraphrase detection}
Now we consider paraphrase detection on the \emph{MSRP} dataset~\cite{dolan2004unsupervised}. On this task, one needs to predict whether or not two sentences are paraphrases. The training set consists of 4076 sentence pairs, and the test set has 1725 pairs. As in~\citet{tai2015improved}, given two sentence representations $\zv_x$ and $\zv_y$, we first compute their element-wise product $\zv_x\odot \zv_y$ and their absolute difference $|\zv_x-\zv_y|$, and then concatenate them together. A logistic regression model is trained on top of the concatenated features to predict whether two sentences are paraphrases. We present our results on the last column of Table~\ref{Table:classification_result}. Our best result is better than the other results that use task-independent methods. 

\begin{table}[t!]
	\centering
	\small
	%\begin{tabular}{lcc|cc}
	\begin{adjustbox}{scale=0.9,tabular=lcc|cc,center}
		\toprule
		& \multicolumn{2}{c|}{\small {Image Annotation}} & \multicolumn{2}{c}{Image Search} \\
		\textbf{Method} & \textbf{R@1}   & \textbf{Med} $r$ & \textbf{R@1}  & \textbf{Med} $r$ \\
		\midrule 
		uni-skip$^\dagger$ & 30.6   & 3 & 22.7   & 4 \\
		bi-skip$^\dagger$ & 32.7   & 3 & 24.2   & 4\\
		combine-skip$^\dagger$ & 33.8   & 3 & 25.9  & 4\\
		\midrule
		\emph{Our Results} & \multicolumn{4}{c}{ } \\
		\midrule
		hierarchical model+emb.  & 32.7  & 3 & 25.3  & 4\\
		composite model+emb.  & 33.8  & 3 & 25.7  & 4\\
		combine+emb.  & {\bf34.4}   & 3 & {\bf26.6}  & 4\\
		\midrule
		\emph{Task-dependent methods} & \multicolumn{4}{c}{ } \\
		\midrule
		DVSA$^*$ & 38.4  & 1 & 27.4  & 3\\
		m-RNN$^\ddagger$ & 41.0  & 2 & 29.0  & 3\\
		\bottomrule
	\end{adjustbox}
	\caption{{Results for image-sentence ranking experiments on the COCO dataset. \textbf{R@K} denotes Recall@K (higher is better) and \textbf{Med} $r$ is the median rank (lower is better). ($\dagger$) taken from \citet{kiros2015skip}. ($*$) taken from \citet{karpathy2015deep}. ($\ddagger$) taken from \citet{mao2014deep}.}}\label{Table:ranking}
	%\vspace{-4mm}
	%\end{tabular}
\end{table}

\paragraph{Image-sentence ranking}
We consider the task of image-sentence ranking, which aims to retrieve items in one modality given a query from the other. We use the COCO dataset~\cite{lin2014microsoft}, which contains 123,287 images each with 5 captions. For development and testing we use the same splits as~\citet{karpathy2015deep}. The development and test sets each contain 5000 images. We further split them into 5 random sets of 1000 images, and report the average performance over the 5 splits. Performance is evaluated using Recall@K, which measures the average times a correct item is found within the top-K retrieved results. We also report the median rank of the closest ground truth result in the ranked list. 

\begin{table}[t]
	\centering
	\small
	\begin{tabular}{lccc}
		\toprule
		\textbf{Method} & $r$ & $\rho$ & \textbf{MSE}  \\
		\midrule 
		uni-skip$^\dagger$  & 0.8477 & 0.7780 & 0.2872 \\
		bi-skip$^\dagger$  & 0.8405 & 0.7696 & 0.2995 \\
		combine-skip$^\dagger$ & 0.8584 & 0.7916 & 0.2687 \\
		\midrule
		\emph{Our Results} & \multicolumn{3}{c}{ } \\
		\midrule
		autoencoder   & 0.8284 & 0.7577 & 0.3258  \\
		future predictor  & 0.8132 & 0.7342 & 0.3450  \\
		hierarchical model  & 0.8333 & 0.7646 & 0.3135  \\
		composite model  & 0.8434 & 0.7767 & 0.2972  \\
		combine  & 0.8533 & 0.7891 & 0.2791 \\
		\midrule
		hierarchical model+emb.  & 0.8352 & 0.7588 & 0.3152  \\
		composite model+emb.  & 0.8500 & 0.7867 & 0.2872 \\
		combine+emb.  & {\bf 0.8618}  & {\bf 0.7983} & {\bf 0.2668}  \\
		\midrule
		\emph{Task-dependent methods} & \multicolumn{3}{c}{ } \\
		\midrule
		Bi-LSTM$^\ddagger$ & 0.8567 & 0.7966 & 0.2736 \\
		Tree-LSTM$^\ddagger$ & 0.8676 & 0.8083 & 0.2532 \\
		\bottomrule
	\end{tabular}
	\caption{{Results on the SICK semantic relatedness task. The evaluation metrics are Pearson's $r$, Spearman's $\rho$ and mean squared error (MSE). ($\dagger$) taken from \citet{kiros2015skip}. ($\ddagger$) taken from \citet{tai2015improved}.}}\label{Table:sick}
	%\vspace{-3.5mm}
\end{table}
We represent images using 4096-dimensional feature vectors from VggNet~\cite{simonyan2014very}. Each caption is encoded using our trained CNN encoder. The training objective is the same pairwise ranking loss as used in~\citet{kiros2015skip}, which takes the form of $ \max(0,\alpha-f(x_n, y_n) + f(x_n, y_m))$, where $f(\cdot,\cdot)$ is the image-sentence score. $(x_n, y_n)$ denotes the related image-sentence pair, and $(x_n, y_m)$ is the randomly sampled unrelated image-sentence pair with $n\not=m$. For image retrieval from sentences, $x$ denotes the caption, $y$ denotes the image, and {\em vice versa}. The objective is to force the matching score of the related pair $(x_n, y_n)$ to be greater than the unrelated pair $(x_n, y_m)$ by a margin $\alpha$, which is set to 0.1 in our experiments. 
%The detailed setup is provided in the Supplementary Material. 

Table~\ref{Table:ranking} shows our results.  Consistent with previous experiments, we empirically found that the encoder trained using the fixed word embedding performed better on this task, hence only results using this method are reported. As can be seen, we obtain the same median rank as in~\citet{kiros2015skip}, indicating that our encoder is as competitive as the skip-thought vectors~\cite{kiros2015skip}. The performance gain between our encoder and the combine-skip model of~\citet{kiros2015skip} on the R@1 score is significant, which shows that the CNN encoder has more discriminative power on retrieving the most correct item than the skip-thought vector. 

\paragraph{Semantic relatedness}
For our final experiment, we consider the task of semantic relatedness on the \emph{SICK} dataset~\cite{marelli2014semeval}, consisting of 9927 sentence pairs. Given two sentences, our goal is to produce a real-valued score between $[1,5]$ to indicate how semantically related two sentences are, based on human generated scores. We compute a feature vector representing the pair of sentences in the same way as on the MSRP dataset.  We follow the method in~\citet{tai2015improved}, and use the cross-entropy loss for training. 
Results are summarized in Table~\ref{Table:sick}. Our result is better than the combine-skip model of~\citet{kiros2015skip}. This suggests that CNN also provides competitive performance at matching human relatedness judgements.

\section{Conclusion}
We presented a new class of CNN-LSTM encoder-decoder models to learn sentence representations from unlabeled text. Our trained convolutional encoder is highly generic, and can be an alternative to the skip-thought vectors of~\citet{kiros2015skip}. Compelling experimental results on several tasks demonstrated the advantages of our approach. 
In future work, we aim to use more advanced CNN architectures~\cite{conneau2016very} for learning generic sentence embeddings. 

\section*{Acknowledgments}
This research was supported by ARO, DARPA, DOE, NGA, ONR and NSF.

\bibliography{emnlp2017}

\begin{thebibliography}{56}
\expandafter\ifx\csname natexlab\endcsname\relax\def\natexlab#1{#1}\fi

\bibitem[{Arora et~al.(2017)Arora, Liang, and Ma}]{Arora2017simple}
Sanjeev Arora, Yingyu Liang, and Tengyu Ma. 2017.
\newblock A simple but tough-to-beat baseline for sentence embeddings.
\newblock In \emph{ICLR}.

\bibitem[{Bastien et~al.(2012)Bastien, Lamblin, Pascanu, Bergstra, Goodfellow,
  Bergeron, Bouchard, Warde-Farley, and Bengio}]{bastien2012theano}
Fr{\'e}d{\'e}ric Bastien, Pascal Lamblin, Razvan Pascanu, James Bergstra, Ian
  Goodfellow, Arnaud Bergeron, Nicolas Bouchard, David Warde-Farley, and Yoshua
  Bengio. 2012.
\newblock Theano: new features and speed improvements.
\newblock \emph{arXiv:1211.5590}.

\bibitem[{Cho et~al.(2014)Cho, Van~Merri{\"e}nboer, Gulcehre, Bahdanau,
  Bougares, Schwenk, and Bengio}]{cho2014learning}
Kyunghyun Cho, Bart Van~Merri{\"e}nboer, Caglar Gulcehre, Dzmitry Bahdanau,
  Fethi Bougares, Holger Schwenk, and Yoshua Bengio. 2014.
\newblock Learning phrase representations using rnn encoder-decoder for
  statistical machine translation.
\newblock In \emph{EMNLP}.

\bibitem[{Collobert et~al.(2011)Collobert, Weston, Bottou, Karlen, Kavukcuoglu,
  and Kuksa}]{collobert2011natural}
Ronan Collobert, Jason Weston, L{\'e}on Bottou, Michael Karlen, Koray
  Kavukcuoglu, and Pavel Kuksa. 2011.
\newblock Natural language processing (almost) from scratch.
\newblock In \emph{JMLR}.

\bibitem[{Conneau et~al.(2016)Conneau, Schwenk, Barrault, and
  Lecun}]{conneau2016very}
Alexis Conneau, Holger Schwenk, Lo{\"\i}c Barrault, and Yann Lecun. 2016.
\newblock Very deep convolutional networks for natural language processing.
\newblock \emph{arXiv:1606.01781}.

\bibitem[{Dai and Le(2015)}]{dai2015semi}
Andrew~M Dai and Quoc~V Le. 2015.
\newblock Semi-supervised sequence learning.
\newblock In \emph{NIPS}.

\bibitem[{Dolan et~al.(2004)Dolan, Quirk, and Brockett}]{dolan2004unsupervised}
Bill Dolan, Chris Quirk, and Chris Brockett. 2004.
\newblock Unsupervised construction of large paraphrase corpora: Exploiting
  massively parallel news sources.
\newblock In \emph{COLING}.

\bibitem[{Gan et~al.(2017)Gan, Singh, Joshi, He, Chen, Gao, and
  Deng}]{gan2017character}
Zhe Gan, PD~Singh, Ameet Joshi, Xiaodong He, Jianshu Chen, Jianfeng Gao, and
  Li~Deng. 2017.
\newblock Character-level deep conflation for business data analytics.
\newblock \emph{arXiv preprint arXiv:1702.02640}.

\bibitem[{Gehring et~al.(2016)Gehring, Auli, Grangier, and
  Dauphin}]{gehring2016convolutional}
Jonas Gehring, Michael Auli, David Grangier, and Yann~N Dauphin. 2016.
\newblock A convolutional encoder model for neural machine translation.
\newblock \emph{arXiv:1611.02344}.

\bibitem[{Goodfellow et~al.(2016)Goodfellow, Bengio, and
  Courville}]{Goodfellow2016Book}
Ian Goodfellow, Yoshua Bengio, and Aaron Courville. 2016.
\newblock \emph{Deep Learning}.
\newblock MIT Press.

\bibitem[{He et~al.(2015)He, Gimpel, and Lin}]{he2015multi}
Hua He, Kevin Gimpel, and Jimmy~J Lin. 2015.
\newblock Multi-perspective sentence similarity modeling with convolutional
  neural networks.
\newblock In \emph{EMNLP}.

\bibitem[{Hill et~al.(2016)Hill, Cho, and Korhonen}]{hill2016learning}
Felix Hill, Kyunghyun Cho, and Anna Korhonen. 2016.
\newblock Learning distributed representations of sentences from unlabelled
  data.
\newblock In \emph{NAACL}.

\bibitem[{Hochreiter and Schmidhuber(1997)}]{hochreiter1997long}
Sepp Hochreiter and J{\"u}rgen Schmidhuber. 1997.
\newblock Long short-term memory.
\newblock In \emph{Neural computation}.

\bibitem[{Hu et~al.(2014)Hu, Lu, Li, and Chen}]{hu2014convolutional}
Baotian Hu, Zhengdong Lu, Hang Li, and Qingcai Chen. 2014.
\newblock Convolutional neural network architectures for matching natural
  language sentences.
\newblock In \emph{NIPS}.

\bibitem[{Hu and Liu(2004)}]{hu2004mining}
Minqing Hu and Bing Liu. 2004.
\newblock Mining and summarizing customer reviews.
\newblock In \emph{SIGKDD}.

\bibitem[{Huang et~al.(2013)Huang, He, Gao, Deng, Acero, and
  Heck}]{huang2013learning}
Po-Sen Huang, Xiaodong He, Jianfeng Gao, Li~Deng, Alex Acero, and Larry Heck.
  2013.
\newblock Learning deep structured semantic models for web search using
  clickthrough data.
\newblock In \emph{CIKM}.

\bibitem[{Iyyer et~al.(2015)Iyyer, Manjunatha, Boyd-Graber, and
  Daum{\'e}~III}]{iyyer2015deep}
Mohit Iyyer, Varun Manjunatha, Jordan~L Boyd-Graber, and Hal Daum{\'e}~III.
  2015.
\newblock Deep unordered composition rivals syntactic methods for text
  classification.
\newblock In \emph{ACL}.

\bibitem[{Johnson and Zhang(2015)}]{johnson2014effective}
Rie Johnson and Tong Zhang. 2015.
\newblock Effective use of word order for text categorization with
  convolutional neural networks.
\newblock In \emph{NAACL HLT}.

\bibitem[{Kalchbrenner and Blunsom(2013)}]{kalchbrenner2013recurrent}
Nal Kalchbrenner and Phil Blunsom. 2013.
\newblock Recurrent continuous translation models.
\newblock In \emph{EMNLP}.

\bibitem[{Kalchbrenner et~al.(2014)Kalchbrenner, Grefenstette, and
  Blunsom}]{kalchbrenner2014convolutional}
Nal Kalchbrenner, Edward Grefenstette, and Phil Blunsom. 2014.
\newblock A convolutional neural network for modelling sentences.
\newblock In \emph{ACL}.

\bibitem[{Karpathy and Fei-Fei(2015)}]{karpathy2015deep}
Andrej Karpathy and Li~Fei-Fei. 2015.
\newblock Deep visual-semantic alignments for generating image descriptions.
\newblock In \emph{CVPR}.

\bibitem[{Kim(2014)}]{kim2014convolutional}
Yoon Kim. 2014.
\newblock Convolutional neural networks for sentence classification.
\newblock In \emph{EMNLP}.

\bibitem[{Kingma and Ba(2015)}]{kingma2014adam}
Diederik Kingma and Jimmy Ba. 2015.
\newblock Adam: A method for stochastic optimization.
\newblock In \emph{ICLR}.

\bibitem[{Kiros et~al.(2015)Kiros, Zhu, Salakhutdinov, Zemel, Urtasun,
  Torralba, and Fidler}]{kiros2015skip}
Ryan Kiros, Yukun Zhu, Ruslan~R Salakhutdinov, Richard Zemel, Raquel Urtasun,
  Antonio Torralba, and Sanja Fidler. 2015.
\newblock Skip-thought vectors.
\newblock In \emph{NIPS}.

\bibitem[{Le and Mikolov(2014)}]{le2014distributed}
Quoc Le and Tomas Mikolov. 2014.
\newblock Distributed representations of sentences and documents.
\newblock In \emph{ICML}.

\bibitem[{Li et~al.(2015)Li, Luong, and Jurafsky}]{li2015hierarchical}
Jiwei Li, Minh-Thang Luong, and Dan Jurafsky. 2015.
\newblock A hierarchical neural autoencoder for paragraphs and documents.
\newblock In \emph{ACL}.

\bibitem[{Li and Roth(2002)}]{li2002learning}
Xin Li and Dan Roth. 2002.
\newblock Learning question classifiers.
\newblock In \emph{ACL}.

\bibitem[{Lin et~al.(2015)Lin, Liu, Yang, Li, Zhou, and
  Li}]{lin2015hierarchical}
Rui Lin, Shujie Liu, Muyun Yang, Mu~Li, Ming Zhou, and Sheng Li. 2015.
\newblock Hierarchical recurrent neural network for document modeling.
\newblock In \emph{EMNLP}.

\bibitem[{Lin et~al.(2014)Lin, Maire, Belongie, Hays, Perona, Ramanan,
  Doll{\'a}r, and Zitnick}]{lin2014microsoft}
Tsung-Yi Lin, Michael Maire, Serge Belongie, James Hays, Pietro Perona, Deva
  Ramanan, Piotr Doll{\'a}r, and C~Lawrence Zitnick. 2014.
\newblock Microsoft coco: Common objects in context.
\newblock In \emph{ECCV}.

\bibitem[{Lin et~al.(2017)Lin, Feng, Santos, Yu, Xiang, Zhou, and
  Bengio}]{lin2017structured}
Zhouhan Lin, Minwei Feng, Cicero Nogueira~dos Santos, Mo~Yu, Bing Xiang, Bowen
  Zhou, and Yoshua Bengio. 2017.
\newblock A structured self-attentive sentence embedding.
\newblock In \emph{ICLR}.

\bibitem[{Mao et~al.(2015)Mao, Xu, Yang, Wang, Huang, and Yuille}]{mao2014deep}
Junhua Mao, Wei Xu, Yi~Yang, Jiang Wang, Zhiheng Huang, and Alan Yuille. 2015.
\newblock Deep captioning with multimodal recurrent neural networks (m-rnn).
\newblock In \emph{ICLR}.

\bibitem[{Marelli et~al.(2014)Marelli, Bentivogli, Baroni, Bernardi, Menini,
  and Zamparelli}]{marelli2014semeval}
Marco Marelli, Luisa Bentivogli, Marco Baroni, Raffaella Bernardi, Stefano
  Menini, and Roberto Zamparelli. 2014.
\newblock Semeval-2014 task 1: Evaluation of compositional distributional
  semantic models on full sentences through semantic relatedness and textual
  entailment.
\newblock \emph{SemEval-2014}.

\bibitem[{Meng et~al.(2015)Meng, Lu, Wang, Li, Jiang, and
  Liu}]{meng2015encoding}
Fandong Meng, Zhengdong Lu, Mingxuan Wang, Hang Li, Wenbin Jiang, and Qun Liu.
  2015.
\newblock Encoding source language with convolutional neural network for
  machine translation.
\newblock In \emph{ACL}.

\bibitem[{Mikolov et~al.(2010)Mikolov, Karafi{\'a}t, Burget, Cernock{\`y}, and
  Khudanpur}]{mikolov2010recurrent}
Tomas Mikolov, Martin Karafi{\'a}t, Lukas Burget, Jan Cernock{\`y}, and Sanjeev
  Khudanpur. 2010.
\newblock Recurrent neural network based language model.
\newblock In \emph{INTERSPEECH}.

\bibitem[{Mikolov et~al.(2013)Mikolov, Sutskever, Chen, Corrado, and
  Dean}]{mikolov2013distributed}
Tomas Mikolov, Ilya Sutskever, Kai Chen, Greg~S Corrado, and Jeff Dean. 2013.
\newblock Distributed representations of words and phrases and their
  compositionality.
\newblock In \emph{NIPS}.

\bibitem[{Mitchell and Lapata(2010)}]{mitchell2010composition}
Jeff Mitchell and Mirella Lapata. 2010.
\newblock Composition in distributional models of semantics.
\newblock \emph{Cognitive science}.

\bibitem[{Pagliardini et~al.(2017)Pagliardini, Gupta, and
  Jaggi}]{Pagliardini2017unsupervised}
Matteo Pagliardini, Prakhar Gupta, and Martin Jaggi. 2017.
\newblock Unsupervised learning of sentence embeddings using compositional
  n-gram features.
\newblock \emph{arXiv:1703.02507}.

\bibitem[{Pang and Lee(2004)}]{pang2004sentimental}
Bo~Pang and Lillian Lee. 2004.
\newblock A sentimental education: Sentiment analysis using subjectivity
  summarization based on minimum cuts.
\newblock In \emph{ACL}.

\bibitem[{Pang and Lee(2005)}]{pang2005seeing}
Bo~Pang and Lillian Lee. 2005.
\newblock Seeing stars: Exploiting class relationships for sentiment
  categorization with respect to rating scales.
\newblock In \emph{ACL}.

\bibitem[{Shen et~al.(2014)Shen, He, Gao, Deng, and Mesnil}]{shen2014latent}
Yelong Shen, Xiaodong He, Jianfeng Gao, Li~Deng, and Gr{\'e}goire Mesnil. 2014.
\newblock A latent semantic model with convolutional-pooling structure for
  information retrieval.
\newblock In \emph{CIKM}.

\bibitem[{Simonyan and Zisserman(2015)}]{simonyan2014very}
Karen Simonyan and Andrew Zisserman. 2015.
\newblock Very deep convolutional networks for large-scale image recognition.
\newblock In \emph{ICLR}.

\bibitem[{Socher et~al.(2011)Socher, Huang, Pennington, Ng, and
  Manning}]{socher2011dynamic}
Richard Socher, Eric~H Huang, Jeffrey Pennington, Andrew~Y Ng, and
  Christopher~D Manning. 2011.
\newblock Dynamic pooling and unfolding recursive autoencoders for paraphrase
  detection.
\newblock In \emph{NIPS}.

\bibitem[{Socher et~al.(2013)Socher, Perelygin, Wu, Chuang, Manning, Ng, and
  Potts}]{socher2013recursive}
Richard Socher, Alex Perelygin, Jean~Y Wu, Jason Chuang, Christopher~D Manning,
  Andrew~Y Ng, and Christopher Potts. 2013.
\newblock Recursive deep models for semantic compositionality over a sentiment
  treebank.
\newblock In \emph{EMNLP}.

\bibitem[{Sordoni et~al.(2015)Sordoni, Bengio, Vahabi, Lioma, Grue~Simonsen,
  and Nie}]{sordoni2015hierarchical}
Alessandro Sordoni, Yoshua Bengio, Hossein Vahabi, Christina Lioma, Jakob
  Grue~Simonsen, and Jian-Yun Nie. 2015.
\newblock A hierarchical recurrent encoder-decoder for generative context-aware
  query suggestion.
\newblock In \emph{CIKM}.

\bibitem[{Srivastava et~al.(2014)Srivastava, Hinton, Krizhevsky, Sutskever, and
  Salakhutdinov}]{srivastava2014dropout}
Nitish Srivastava, Geoffrey~E Hinton, Alex Krizhevsky, Ilya Sutskever, and
  Ruslan Salakhutdinov. 2014.
\newblock Dropout: a simple way to prevent neural networks from overfitting.
\newblock \emph{JMLR}.

\bibitem[{Srivastava et~al.(2015)Srivastava, Mansimov, and
  Salakhudinov}]{srivastava2015unsupervised}
Nitish Srivastava, Elman Mansimov, and Ruslan Salakhudinov. 2015.
\newblock Unsupervised learning of video representations using lstms.
\newblock In \emph{ICML}.

\bibitem[{Sutskever et~al.(2014)Sutskever, Vinyals, and
  Le}]{sutskever2014sequence}
Ilya Sutskever, Oriol Vinyals, and Quoc~V Le. 2014.
\newblock Sequence to sequence learning with neural networks.
\newblock In \emph{NIPS}.

\bibitem[{Tai et~al.(2015)Tai, Socher, and Manning}]{tai2015improved}
Kai~Sheng Tai, Richard Socher, and Christopher~D Manning. 2015.
\newblock Improved semantic representations from tree-structured long
  short-term memory networks.
\newblock In \emph{ACL}.

\bibitem[{Wang and Cho(2016)}]{wanglarger}
Tian Wang and Kyunghyun Cho. 2016.
\newblock Larger-context language modelling with recurrent neural network.
\newblock In \emph{ACL}.

\bibitem[{Wiebe et~al.(2005)Wiebe, Wilson, and Cardie}]{wiebe2005annotating}
Janyce Wiebe, Theresa Wilson, and Claire Cardie. 2005.
\newblock Annotating expressions of opinions and emotions in language.
\newblock In \emph{Language resources and evaluation}.

\bibitem[{Wieting et~al.(2016)Wieting, Bansal, Gimpel, and
  Livescu}]{wieting2015towards}
John Wieting, Mohit Bansal, Kevin Gimpel, and Karen Livescu. 2016.
\newblock Towards universal paraphrastic sentence embeddings.
\newblock In \emph{ICLR}.

\bibitem[{Yin and Sch{\"u}tze(2015)}]{yin2015convolutional}
Wenpeng Yin and Hinrich Sch{\"u}tze. 2015.
\newblock Convolutional neural network for paraphrase identification.
\newblock In \emph{HLT-NAACL}.

\bibitem[{Yu and Dredze(2015)}]{yu2015learning}
Mo~Yu and Mark Dredze. 2015.
\newblock Learning composition models for phrase embeddings.
\newblock \emph{TACL}.

\bibitem[{Zhang et~al.(2015)Zhang, Zhao, and LeCun}]{zhang2015character}
Xiang Zhang, Junbo Zhao, and Yann LeCun. 2015.
\newblock Character-level convolutional networks for text classification.
\newblock In \emph{NIPS}.

\bibitem[{Zhao et~al.(2015)Zhao, Lu, and Poupart}]{zhao2015self}
Han Zhao, Zhengdong Lu, and Pascal Poupart. 2015.
\newblock Self-adaptive hierarchical sentence model.
\newblock \emph{arXiv:1504.05070}.

\bibitem[{Zhu et~al.(2015)Zhu, Kiros, Zemel, Salakhutdinov, Urtasun, Torralba,
  and Fidler}]{zhu2015aligning}
Yukun Zhu, Ryan Kiros, Rich Zemel, Ruslan Salakhutdinov, Raquel Urtasun,
  Antonio Torralba, and Sanja Fidler. 2015.
\newblock Aligning books and movies: Towards story-like visual explanations by
  watching movies and reading books.
\newblock In \emph{ICCV}.

\end{thebibliography}
\bibliographystyle{emnlp_natbib}

\end{document}